\documentclass[draftclsnofoot,onecolumn]{IEEEtran}
\ifCLASSINFOpdf
\else
\fi
\usepackage{amsmath}
\usepackage{amsfonts,amssymb}
\usepackage{graphicx}
\usepackage{subfigure}
\usepackage{color}
\usepackage{bm}

\begin{document}
%
\title{Unsupervised Feature Selection via Multi-step Markov Transition Probability}
%
%
%

\author{Yan~Min, Mao~Ye$^*$, ~\IEEEmembership{Member,~IEEE,}
Liang~Tian, Yulin Jian,
Ce~Zhu, ~\IEEEmembership{Fellow,~IEEE,}
and Shangming~Yang

\thanks{Yan Min, Mao Ye, Liang Tian and Yulin Jian are with the School of Computer Science and Engineering, University of Electronic Science and Technology of China, Chengdu 611731, P.R. China (e-mail: cvlab.uestc@gmail.com).}
\thanks{Ce Zhu is with the School of Information and Communication Engineering, University of Electronic Science and Technology of China, Chengdu 611731, P.R. China (e-mail: eczhu@uestc.edu.cn)}
\thanks{Shangming Yang is with the School of Information and Software Engineering, University of Electronic Science and Technology of China, Chengdu 610054, P.R. China}
\thanks{*corresponding author}}

\maketitle

\begin{abstract}
Feature selection is a widely used dimension reduction technique to select feature subsets because of its interpretability. Many methods have been proposed and achieved good results, in which the relationships between adjacent data points are mainly concerned.  But the possible associations between  data pairs that are may not adjacent are always neglected. Different from previous methods, we propose a novel and very simple approach for unsupervised feature selection, named MMFS (Multi-step Markov transition probability for Feature Selection).  The idea is using multi-step Markov transition probability to describe the relation between any data pair. Two ways from  the positive and negative viewpoints are employed respectively to keep the data structure after feature selection. From the positive viewpoint, the maximum transition probability that can be reached in a certain number of steps is used to describe the relation between two points. Then, the features which can keep the compact data structure are selected. From the viewpoint of negative, the minimum transition probability that can be reached in a certain number of steps is used to describe the relation between two points. On the contrary, the features that least maintain the loose data structure are selected. And the two ways can also be combined. Thus three algorithms are proposed.  Our main contributions are  a novel feature section approach which uses multi-step transition probability to characterize the data structure, and three algorithms proposed from the positive and negative aspects for keeping data structure.
The performance of our approach is compared with the state-of-the-art methods on eight real-world data sets, and the experimental results show that the proposed MMFS is effective in unsupervised feature selection.
\end{abstract}


%
\IEEEpeerreviewmaketitle

\section{Introduction}
%
%
%
%
\IEEEPARstart{I}{n} machine learning, computer vision, data mining, and other fields, with the increase of the difficulty of research objects, we have to deal with some high-dimensional data, such as text data, image data, and various gene expression data. When processing high-dimensional data, we are confronted with the curse of dimensionality and some other difficulties caused by too many features, which not only makes the prediction results inaccurate, but also consumes the calculation time. So it is very important to process high dimensional data. There are three approaches to deal with high-dimensional data, i.e., feature extraction, feature compression and feature selection. Feature extraction generally maps data from high-dimensional  space to low-dimensional space, such as principal component analysis (PCA), linear discriminant analysis (LDA) and neural networks.  Feature compression compresses the original feature value to 0 or 1 by quantization. Feature selection uses some evaluation criteria to select feature subsets from the original feature set. In general, the features selected by feature selection method are more interpretable. In addition, the discrimination ability of the selected features is not weaker than that of the extracted or compressed features. As an effective mean to remove irrelevant features from high-dimensional data without reducing performance, feature selection has attracted many attentions in recent years.
\par According to whether it is independent of the classification algorithm used later, the feature selection method can be categorized as filter, wrapper, and embedded methods \cite{wang2016sparse}. Filter method \cite{liu2013global, MingkuiMinimax} first selects features and then the feature subset obtained by filtering operation is used to train the classifier, so filter method is fully independent of the classifier. The essence of this method is to use some indexes in mathematical statistics to rank feature subsets, some popular evaluation criteria of filter methods include fisher score, correlation coefficient, mutual information, information entropy, and variance threshold. According to the objective function, wrapper method \cite{freeman2013feature}, \cite{xue2012particle} selects or excludes several features at a time until the best subset is selected. In other words, wrapper method wraps the classifier and feature selection into a black box, and evaluates the performance of the selected feature according to its accuracy on the feature subset. Embedded methods \cite{nie2014clustering}, \cite{mohsenzadeh2013relevance} get the weight coefficients of each feature by learning, and rank features according to the coefficients. The difference between the embedded and filter methods is whether determines the selection of features through training.

\begin{figure*}[t]	
	\centering
	\includegraphics[width=16cm]{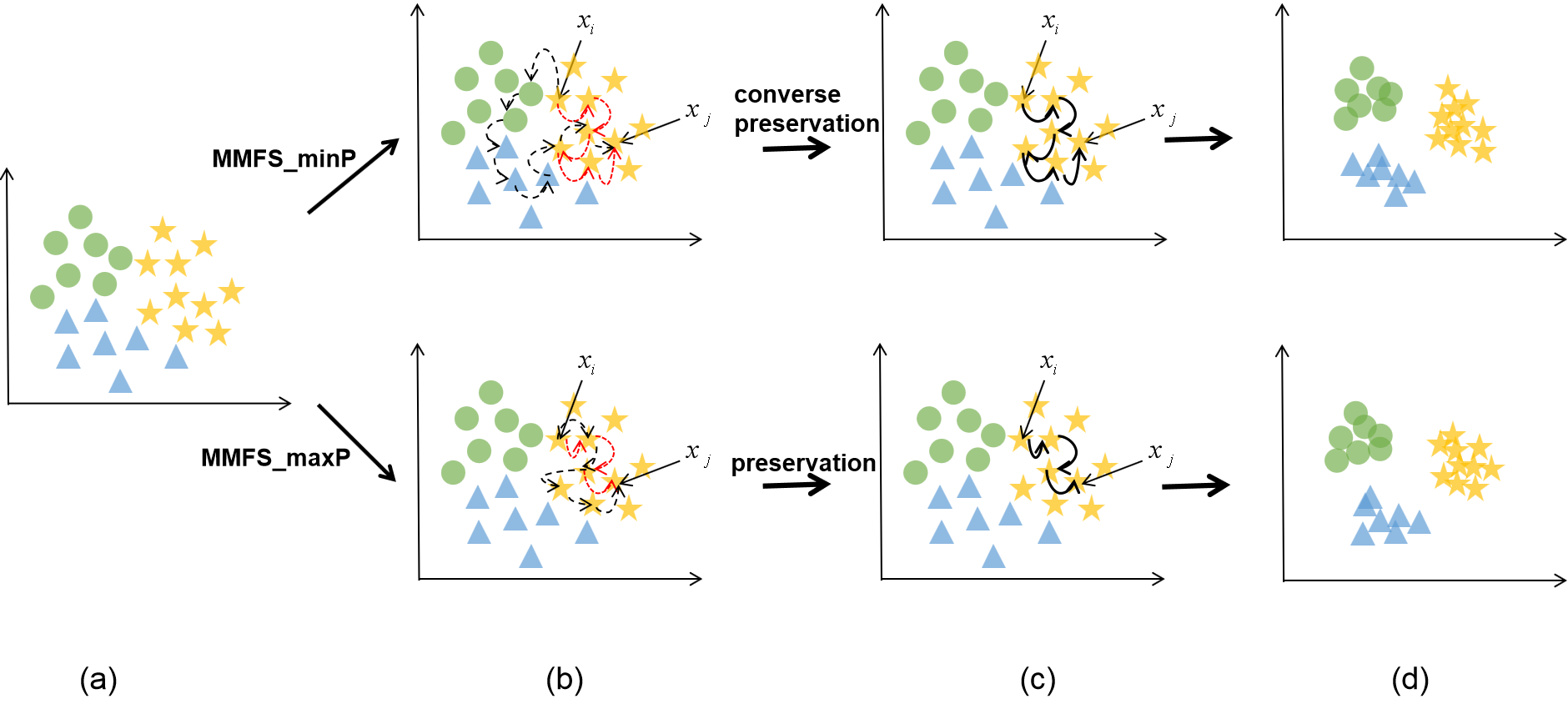}
	\caption{High dimensional data (a) can preserve structure by multi-step Markov transition probability, through the corresponding minimum transition probability to connect more points or the maximum transition probability to fewer points to obtain the relationship between data points (b), and different structure preservation algorithms will get different relationship  between data points (c). For the negative side (MMFS\_minP), the path connecting most data points will be discarded; while for the positive side (MMFS\_maxP), the path connecting minimum data points will be kept. After feature selection, these two algorithms can not only keep the intrinsic data structure, but also shorten the distance between data points in the same class (d) by the obtained relationships.}
	\label{Fig.main}
\end{figure*}

\par  Since any tiny part of manifold can be regarded as Euclidean space, assuming that the data is a low dimensional manifold uniformly sampled in a high dimensional Euclidean space, manifold learning is to recover the low dimensional manifold structure from the high dimensional sampled data, that is, to find the low dimensional manifold in the high dimensional space, and find out the corresponding embedded mapping to achieve dimension reduction or data visualization. Some popular manifold learning methods include isometric feature mapping (ISOMAP) \cite{tenenbaum2000global}, Locally Linear Embedding (LLE) \cite{roweis2000nonlinear} and Laplacian eigenmaps (LE) \cite{belkin2002laplacian}, \cite{belkin2003laplacian}.
From this research line, many unsupervised feature selection methods are proposed, such as \cite{zhao2007spectral}, \cite{he2006laplacian} and \cite{cai2010unsupervised}, which select features by the local structure information of data points, and these methods show that the local structure information is helpful to select the effective feature subset.
However, most of these methods only use the structure information between adjacent data points, and  the manifold structure of the whole data is not sufficiently employed.

\par In this paper, inspired by the above analysis, we propose a new unsupervised feature selection approach called MMFS (Multi-step Markov transition probability for Feature Selection). As shown in Fig. 1, the core idea is to use multi-step transition probability to characterize data relationships on manifold.
Based on the idea of keeping data structure, we do feature selection from both positive and negative aspects. On the positive side, the maximum multi-step transition probability that can be reached in a certain number of steps between any data pair is used to describe the
compact data structure. The features which can better keep this data structure are selected. On the negative side, to characterize the loose data structure, the minimal multi-step transition probability that can be reached in a certain number of steps between any data pair is used. The features that least maintain this loose data structure are selected. These two ways are also be combined to form a new algorithm which can obtain average performance. Thus, three algorithms are proposed.

The main contributions of our work can be summarized as follows.
\par 1) A novel unsupervised feature selection approach is proposed, which can sufficiently use and keep data structure on manifold. Instead of directly using Euclidean distance, multi-step Markov transition probability is used to describe the data structure.
\par 2) Different from the existing solutions, we design two algorithms from both positive and negative viewpoints. Features which better and least keep the corresponding data structures are selected. After feature selection, the data in the low-dimensional space will be more compact.
\par 3) Comprehensive experiments are performed on eight benchmark data sets, which show the good performance of the proposed approach compared with the state-of-the-art unsupervised feature selection methods.
\par The rest of this paper is organized as follows. Section \uppercase\expandafter{\romannumeral2} reviews some related work, and Section \uppercase\expandafter{\romannumeral3} presents the notations and definitions used in this paper. In Section \uppercase\expandafter{\romannumeral4}, we propose the new approach MMFS. The optimization method is presented in Section \uppercase\expandafter{\romannumeral5} and the experimental results are presented and analyzed in Section \uppercase\expandafter{\romannumeral6}. Finally, conclusions are given in Section \uppercase\expandafter{\romannumeral7}.

\section{RELATED WORK}
According to whether to use label information or not, feature selection methods can be divided into three different types: supervised feature selection, semi-supervised feature selection and unsupervised feature selection.

Supervised feature selection methods with ground-truth class labels usually make full use of these ground-truth class labels to select more discriminative features. Most supervised methods valuate feature relevance based on the correlations of the features with the labels. e.g., Fisher Score \cite{gu2012generalized}, Relief-F \cite{robnik2003theoretical}, spectral analysis \cite{zhao2007semi}, trace ratio \cite{nie2008trace}, information entropy \cite{yu2003feature}, Pearson correlation coefficients \cite{lee1988thirteen}, mutual information \cite{peng2005feature}, \cite{yang2012effective}, and Hilbert Schmidt independence criterion \cite{song2007supervised}. Recently, some methods \cite{he20122}, \cite{cai2011multi} apply  $l_{2,1}$-norm to improve the performance.

A number of semi-supervise feature selection methods have been presented during the past ten years. Semi-supervised feature selection focuses on the problem of using a small number of labeled data and a large number of unlabeled data for feature selection. Most semi-supervised feature selection methods score the features based on a ranking criterion, such as Laplacian score, Pearson's correlation coefficient and so on. For example, Zhao et al. \cite{zhao2007semi} presented a semi-supervised feature selection method based on spectral analysis. The method proposed by Doquire et al. \cite{doquire2013graph} introduced a semi-supervised feature selection algorithm that relies on the notion of Laplacian score. And Xu et al. \cite{xu2016semisupervised} combined a max-relevance and min-redundancy criterion based on Pearson's correlation coefficient with semi-supervised feature selections.

\par Unsupervised feature selection without any label information is more difficult and challenging. Because ground-truth class labels are costly to be obtained, it is desirable to develop unsupervised feature selection methods. Our method belongs to this category which will be detailed in the following.
\subsection{Unsupervised Feature Selection Method}

An important approach to achieving unsupervised feature selection is to select features by local structure information of data points. There are many ways to apply the local structure information of data points. In the early stage, the methods usually construct the affinity graph to get the local structure information first, and then select the features. For example, Zhao et al. \cite{zhao2007spectral} proposed a unified framework for feature selection based on spectral graph theory, which is based on general similarity matrix.
\par In the later stage, these methods are able to simultaneously obtain structural information and do feature selection. Gu et al. \cite{gu2012locality} proposed a locality preserving feature selection method; it aims to minimize the locality preserving criterion based on a subset of features by learning a linear transformation. Hou et al.\cite{hou2013joint} defined a novel unsupervised feature selection framework, in which the embedding learning and sparse regression perform jointly. Nie et al. \cite{nie2016unsupervised} came up with an unsupervised feature selection framework, which can simultaneously performs feature selection and local structure learning. Shi et al. \cite{shi2016cluster} incorporated spectral clustering, discriminative analysis, and correlation information between multiple views into a unified framework. Zhao et al. \cite{zhao2015graph} presented an unsupervised feature selection approach, which selects features by preserving the local structure of the original data space via graph regularization and reconstructing each data point approximately via linear combination. An et al. \cite{an2017unsupervised} preserved locality information by a probabilistic neighborhood graph to select features and combined feature selection and robust joint clustering analysis. The method proposed by Fan et al. \cite{fan2017structure} can select more informative features and learn the structure information of data points at the same time.  Dai et al. \cite{dai2018unsupervised} proposed method that each feature is represented by linear combination of other features while maintaining the local geometrical structure and the ordinal locality of original data.
\par Adaptive methods are also introduced to select more effective features. The method proposed by Luo et al. \cite{luo2017adaptive} selects features by an adaptive reconstruction graph to learn local structure information, and imposing a rank constraint on the corresponding Laplacian matrix to learn the multi-cluster structure. Chen et al. \cite{chen2018local} came up with a local adaptive projection framework that can simultaneously learns an adaptive similarity matrix and a projection matrix. Zhang et al. \cite{zhang2019unsupervised} proposed a method that can make the similarities under different measure functions unify adaptively and induced  $l_{2,0}$ constraint to select features.

%
\par In addition, various methods are also used to achieve unsupervised feature selection. Some unsupervised feature selection methods removed redundant features to improve the performance. Li et al. \cite{li2018generalized} removed the redundant features and embedded the local geometric structure of data into the manifold learning to preserve the most discriminative information. Wang et al. \cite{wang2018hierarchical} selected features hierarchically to prune redundant features and preserve useful features. Nie et al. \cite{nie2018general} defined an auto-weighted feature selection framework via global redundancy minimization to select the non-redundant features. Xue et al. \cite{xue2019self} presented a self-adaptive algorithm based on EC method to solve the local optimal stagnation problem caused by a large number of irrelevant features.

\subsection{Markov Walk}
\par Our method MMFS is also closely related to the methods based on Markov random walks. Szummer et al. \cite{szummer2002partially} combined a limited number of labeled samples with Markov random walk representation on unlabeled samples to classify a large number of unlabeled samples. Haeusser et al. \cite{haeusser2017associative} used a two-steps Markov random walk on this graph that starts from the source domain samples and return to the same kind of samples through the target domain data in the embedded space. Sun et al. \cite{sun2019neural} proposed a method, which applying random walker to define the diffusion distance modules for measuring the distance among nodes on graph. Instead of their approaches, we use multi-step transition probability to characterize the data structure and try to select the feature subset which can keep the original data structure.

\section{NOTATIONS AND DEFINITIONS}
We summarize the notations and the definitions used in our paper. We denote all the matrices with boldface uppercase letters and vectors with boldface lowercase letters. For a matrix $\mathbf{M}$, the element in the $ {i_{th}} $ row and $ {j_{th}} $ column is represented as ${m_{ij}}$ , and the transpose of matrix $\mathbf{M}$ is denoted as ${\mathbf{M}^T} $. Its Frobenius norm is denoted by $ ||\mathbf{M}|{|_F} = \sqrt {\sum\nolimits_{i = 1}^n {\sum\nolimits_{j = 1}^m {m_{ij}^2} } } $. The $l_{2,1}$-norm of the matrix $\mathbf{M}$ is denoted as $ ||\mathbf{M}|{|_{2,1}} = \sum\nolimits_{i = 1}^n {\sqrt {\sum\nolimits_{j = 1}^m {m_{ij}^2} } } $.

\begin{table}[]
	 \caption{Notations with Descriptions}
	 \centering
	\begin{tabular}{ll}
		\hline
		Notation                                         & Description                        \\
		\hline
		d                                                & Dimensionality of data point       \\
		N                                                & Number of instances in data matrix \\
		\textbf{X} $\in {\mathbb{R}^{{\rm{d}} \times {\rm{N}}}}$ & Data matrix                        \\
		\textbf{F} $\in {\mathbb{R}^{{\rm{N}} \times {\rm{d}}}}$ & Templates for feature selection \\
		\hline
	\end{tabular}
\end{table}

 \section{The Proposed Approach}

Let $\mathbf{X} = [{{\mathbf{x}}_1}, \cdots, {{\mathbf{x}}_N}]$ denotes a $d$-dimensional data matrix consisting of $N$ instances. In this paper, we want to learn the spatial structure around each point, so that the final selected features can also reflect the spatial structure of the original data in high-dimensional space. In order to obtain the relationship between any two points, we achieve our purpose from  negative and positive sides respectively. Multi-step transition probability is used to characterize the data spatial structure. Assume each data point in the high dimensional space is a node (state). Transition probability refers to the conditional probability that a Markov chain is in one state at a certain time, and then it will reach another state in another certain time. The one-step transition probability $ {\mathbf{P}_{ij}} $ from node $i$ to node $j$ can be defined as follows:
\begin{equation}
	{\mathbf{P}_{ij}} = \displaystyle\frac{{{\mathbf{M}_{ij}}}}{{\sum\nolimits_{m = 1}^N {{\mathbf{M}_{im}}} }}
\end{equation}
where
\begin{equation}
 \displaystyle\textbf{M}\mathop{{}}\nolimits_{{ij}}=\frac{{1}}{{{ \left( {{\left. {\frac{{\textbf{D}\mathop{{}}\nolimits_{{ij}}}}{{{\mathop{ \sum }\nolimits_{{m=1}}^{{N}}{\textbf{D}\mathop{{}}\nolimits_{{im}}}}}}+ \alpha } \right) }}\right. }}}.
\end{equation}
Furthermore, the self transition probability $\textbf{P}_{ii} = 0$ for any $i$ data point. Here, \textbf{D} is a matrix of Euclidean distances between data points and $\alpha$ is a very small constant to avoid the denominator becoming zero. The closer the distance is, the larger the transition probability is.

It is well known that Euclidean distance does not make sense in high dimensional space. Fortunately, any tiny part of manifold can be regarded as Euclidean space, so the one-step transition probability can be computed between the data points that are very close to each other. Naturally, this problem can be converted to select the nearest $k$ points around any data point to calculate the transition probability. So we redefine the one-step probability $ {\mathbf{P}_{ij}}$. That is $\textbf{M}_{ii}$ = 0 and $ \textbf{M}_{ij} $ = 0 when the data point $j$ is not one of the $k$ nearest neighbors of data point $i$. Naturally, the $n$-step transition probability matrix can be calculated as
\begin{equation}
{\textbf{P}^{(n)}} = {\textbf{P}^{(n - 1)}}{\textbf{P}^{(1)}}.
\end{equation}

In the remaining parts, we will try to preserve the original data structure from the positive and negative sides and thus propose three algorithms. First, the algorithm based on the negative viewpoint is illustrated. Then, the algorithm from the positive side is presented. In the end, the third algorithm combing the positive and negative algorithms is proposed.

\subsection{Unsupervised Feature Selection based on Minimum Multi-step Transition Probability}
\par The first algorithm is called MMFS$\_$minP (Multi-step Markov transition probability for feature selection based on minimum probability).
The relationship between a node and any other $n$-step reachable point is described by using the minimum transition probability, i.e., the minimum transition probability at some $t$ step, for $t\leq n$, where $n$ is a parameter. In this way, we can get a matrix $ \mathbf{V_{1}} \in {\mathbb{R}^{{\rm{N}} \times {\rm{N}}}}$  that each element represents the actual distance relationships between each data point and their $n$-step reachable points. Please note that $ {\mathbf{ {V_{1}}}_{ii}} $ = 0. In order to ensure that the sum of each row of the  matrix $\mathbf{V}_1$ is $1$, we perform row-wise normalization over $\mathbf{V_{1}}$.

Based on the minimum reachable relation matrix $V_1$ in $n$ steps, the data relationships are characterized. Thus, we have a template for feature selection as follows,
\begin{equation}
\mathbf{F_1} = \mathbf{V_1}{\mathbf{X}^T}.
\end{equation}
Since we select the $k$ nearest points of each data point to determine the actual distance relationship between the two points, this allows our template naturally maintains the manifold structure. So, we have the following objection,
\begin{equation}
\mathop {\min }\limits_\textbf{w} ||{\textbf{X}^T}\textbf{W} - \mathbf{F_1}||_F^2 + \lambda ||\textbf{W}||_{2,1}^2
\end{equation}
where $\mathbf{W}\in {\mathbb{R}^{{\rm{d}} \times {\rm{d}}}} $ is the weight matrix to make the original data approach to the constructed template $ \mathbf{F_{1}} $.  The first term denotes the error between each weighted sample and the template. The second regularization term is used to force the weight matrix \textbf{W} to be row sparse for feature selection. $ \lambda > $ 0 is a regularization parameter used to balance the first and second terms.

Since the minimum transition probability is used to construct the relationship matrix $ \mathbf{V_1} $, we actually obtain the loose relationship between data points. We should choose those features which do not keep these relationships. Thus, after feature selection, the distance between data points in the same class should be shorten.  So we arrange the row vectors of the optimized \textbf{W} in ascending order according to the value of $l_2$ norm; the first $s$ features are selected.

\begin{table}[t]
	\centering
	\begin{tabular}{p{8cm}}	
		\hline
		\textbf{Algorithm 1} MMFS\_minP                                                                                                  \\
		\hline
		\textbf{Input:} Data matrix \textbf{X}, the parameter $ \lambda $, the steps $n$ and the selected feature number $s$.                               \\
		\textbf{Initialize:} \textbf{Q} as an identity matrix.                                                                               \\
		\textbf{while} not converge \textbf{do}                                                                                             \\
		\quad 1. Update $ {\textbf{W}_{t + 1}} = {\left( {\textbf{X}{\textbf{X}^T} + \lambda \textbf{Q}} \right)^{ - 1}}\textbf{X}\mathbf{F_1} $ .                                                                                                      \\
		\quad 2. Update the diagonal matrix $ {\textbf{Q}_{t + 1}} $ , where the $ {j_{th}} $ diagonal element is  $ \frac{{\sum\nolimits_{i = 1}^d {\sqrt {||{\textbf{W}^i}||_2^2 + \varepsilon } } }}{{\sqrt {||{\textbf{W}^{\rm{j}}}||_2^2 + \varepsilon } }} $                                                 \\
		\quad 3. $t = t + 1$.                                                                                                     \\
		\textbf{	end while}                                                                                                         \\
		\textbf{Output:} Obtain optimal matrix \textbf{W} and calculate each $\|{\textbf{W}^i}|{|_2},i = 1,2,\cdots,d$, then sort in ascending order and select the $s$ top ranking features.\\
		\hline
	\end{tabular}
\end{table}

\subsection{Unsupervised Feature Selection based on Maximum Multi-step Transition Probability}
\par The second algorithm is called MMFS$\_$maxP (Multi-step Markov transition probability for feature selection based on maximum probability).
Instead of the first algorithm,  we use the maximum transition probability to express the relationship between the data point and any other reachable points in $n$ steps, i.e., the corresponding step is recorded as $t$ where $t \leq  n$. As the same as above subsection, we can get a relational matrix $ \mathbf{V_2} $, and normalize the matrix $ \mathbf{V_2} $ by rows. Finally, we get the template as the following,
\begin{equation}
\mathbf{F_2} = \mathbf{V_2}{\textbf{X}^T}.
\end{equation}
Then the objective function of the proposed MMFS\_maxP method is
\begin{equation}
\mathop {\min }\limits_\textbf{w} ||{\textbf{X}^T}\textbf{W} - \mathbf{F_2}||_F^2 + \lambda ||\textbf{W}||_{2,1}^2.
\end{equation}
Different from the method MMFS\_minP, since we use the corresponding maximum transition probability to construct the relationship matrix $\mathbf{V_2} $ which represents a more compact relationship, we should choose features from the positive viewpoint. In order to shorten the distance between data points in the same class, we should select the first $s$ features in descending order according to the value of $l_2$ norm of the optimized matrix \textbf{W}.

\subsection{The Combination of Two Algorithms }
In order to combine the above two algorithms, we propose an algorithm called MMFS\_inter. We first find the intersection of the features selected by MMFS\_minP and MMFS\_maxP, and set the number of features in the intersection as $s_1$. Suppose the number of features required to be selected is $s$, when the number of features from the intersection is not enough, we select the first $(s-s_1)/2 $ features from the feature sequences selected by the above two algorithms.

\begin{table}[t]
	\centering
	\begin{tabular}{p{8cm}}	
		\hline
		\textbf{Algorithm 2} MMFS\_maxP                                                                                                  \\
		\hline
		\textbf{Input:} Data matrix \textbf{X}, the parameter $ \lambda $, the steps $n$ and the selected feature number $s$.                               \\
		\textbf{Initialize:} \textbf{Q} as an identity matrix.                                                                               \\
		\textbf{while} not converge \textbf{do}                                                                                             \\
		\quad 1. Update $ {\textbf{W}_{t + 1}} = {\left( {\textbf{X}{\textbf{X}^T} + \lambda \textbf{Q}} \right)^{ - 1}}\textbf{X}\mathbf{F_2} $ .                                                                                                      \\
		\quad 2. Update the diagonal matrix $ {\textbf{Q}_{t + 1}} $, where the $j$th diagonal element is  $ \frac{{\sum\nolimits_{i = 1}^d {\sqrt {||{\textbf{W}^i}||_2^2 + \varepsilon } } }}{{\sqrt {||{\textbf{W}^{\rm{j}}}||_2^2 + \varepsilon } }} $                                                 \\
		\quad 3. $t = t + 1$.                                                                                                     \\
		\textbf{	end while}                                                                                                         \\
		\textbf{Output:} Obtain optimal matrix \textbf{W} and calculate each $\|{\textbf{W}^i}|{|_2},i = 1,2,\cdots,d  $, then sort in descending order and select the $s$ top ranking features.\\
		\hline
	\end{tabular}
\end{table}

\section{optimization}
In this section, a common model is applied to describe the optimization steps for the above two objectives:
\begin{equation}
\mathop {\min }\limits_\textbf{w} ||{\textbf{X}^T}\textbf{W} - \textbf{F}||_F^2 + \lambda ||\textbf{W}||_{2,1}^2.
\end{equation}
Obviously,  $ ||\textbf{W}||_{2,1} $ can be zero in theory, but this will cause the problem (8) to be non-differentiable. To avoid this situation, $ ||\textbf{W}||_{2,1}^2 $ is  regularized as $ {\left( {\sum\nolimits_{j = 1}^d {\sqrt {||{\textbf{W}^{\rm{j}}}||_2^2 + \varepsilon } } } \right)^2} $  where $ \varepsilon  $  is a small enough constant. It follows that
\begin{equation}
\mathop {\min }\limits_\textbf{w} \left( {||{\textbf{X}^T}\textbf{W} - \textbf{F}||_F^2 + \lambda {{\left( {\sum\nolimits_{j = 1}^d {\sqrt {||{\textbf{W}^{\rm{j}}}||_2^2 + \varepsilon } } } \right)}^2}} \right).
\end{equation}
This problem is equal to problem (8) when $ \varepsilon  $ is infinitely close to zero.
Assume the function
\begin{equation}
L(\textbf{W}) = ||{\textbf{X}^T}\textbf{W} - \textbf{F}||_F^2 + \lambda {\left( {\sum\nolimits_{j = 1}^d {\sqrt {||{\textbf{W}^{\rm{j}}}||_2^2 + \varepsilon } } } \right)^2}.
\end{equation}
\par Finding the partial derivative of L (\textbf{W}) with respect to \textbf{W}, and setting it to zero, we have
\begin{equation}
\frac{{\partial L(\textbf{W})}}{{\partial \textbf{W}}} = 2\textbf{X}({\textbf{X}^T}\textbf{W} - \textbf{F}) + 2\lambda \textbf{Q}\textbf{W} = 0,
\end{equation}
where \textbf{Q} is a diagonal matrix whose  $ {j_{th}}$  diagonal element is
\begin{equation}
{q_{jj}} = \frac{{\sum\nolimits_{i = 1}^d {\sqrt {||{\textbf{W}^i}||_2^2 + \varepsilon } } }}{{\sqrt {||{\textbf{W}^{\rm{j}}}||_2^2 + \varepsilon } }}.
\end{equation}
The matrix \textbf{Q} is unknown and depends on \textbf{W}, we solve \textbf{Q} and \textbf{W} iteratively. With the matrix \textbf{W} fixed, \textbf{Q} can be obtained by Eq. (12). When the matrix \textbf{Q} fixed, according to Eq. (11), we can get $\mathbf{W}$ as follows,
\begin{equation}
\textbf{W} = {\left( {\textbf{X}{\textbf{X}^T} + \lambda \textbf{Q}} \right)^{ - 1}}\textbf{X}\textbf{F}.
\end{equation}

We presents an iterative algorithm to find the optimal solution of \textbf{W}. In each iteration, \textbf{W} is calculated with the current \textbf{Q}, and the \textbf{Q} is updated according to the current calculated \textbf{W}. The iteration procedure is repeated until the algorithm converges.
\par Based on the above analysis, we summarize the whole procedure for solving problem (5) in Algorithm 1 and the procedure for solving problem (7) in Algorithm 2.

\begin{table}[t]
	\caption{Descriptions of Data Sets}
	\centering
	\begin{tabular}{llllll}
		\hline
		ID &Data set &\# Instances &\# Features &\# Classes  &Data type\\
		\hline
		1  & Isolet1  & 1560       & 617       & 26       &Speech data\\
		2  & COIL20   & 1440       & 1024      & 20       &Object image\\
		3  & TOX-171  & 171        & 5748      & 4        &Microarray\\
		4  & Lung     & 203        & 3312      & 5        &Microarray\\
		5  & AT\&T     & 400        & 644       & 40      &Face image\\
		6  & ORL10P   & 100        & 10304     & 10      &Face image\\
		7  & YaleB    & 2414       & 1024      & 38      &Face image\\
		8  & USPS     & 9298       & 256       & 10     &Digit\\
		\hline
	\end{tabular}
\end{table}

\begin{table*}[t]
	\caption{Accuracy (acc\%$  \pm  $std\%) of different unsupervised feature selection methods. \protect\\
		The best result are enlighten in bold and the second best is underlined.}
	\centering
	\begin{tabular}{lllllllll}
		\hline
		Datasets & Isolet1    & COIL20     & AT\&T       & YaleB      & USPS       & ORL10P     & Lung       & TOX\-171    \\
		\hline
		All Features & 58.21 $ \pm $ 3.48 & 58.77 $ \pm $ 5.05 & 60.84 $ \pm $ 3.63 & 9.61 $ \pm $ 0.64  & 65.63 $ \pm $ 2.75 & 67.04 $ \pm $ 7.92 & 69.23 $ \pm $ 9.11 & 42.16 $ \pm $ 2.63 \\
		LS           & 59.36 $ \pm $ 3.60 & 54.97 $ \pm $ 4.39 & 59.71 $ \pm $ 3.28 & 9.98 $ \pm $ 0.48  & 62.49 $ \pm $ 4.16 & 61.91 $ \pm $ 6.86 & 60.44 $ \pm $ 9.26 & 39.60 $ \pm $ 4.38 \\
		MCFS         & 60.74 $ \pm $ 3.88 & 58.93 $ \pm $ 4.76 & 61.13 $ \pm $ 3.62 & 9.73 $ \pm $ 0.71  & 64.96 $ \pm $ 4.04 & 68.31 $ \pm $ 8.03 & 64.66 $ \pm $ 7.93 & 40.58 $ \pm $ 3.07 \\
		UDFS         & 57.92 $ \pm $ 3.33 & 59.79 $ \pm $ 4.34 & 60.75 $ \pm $ 3.40 & 11.36 $ \pm $ 0.62 & 63.46 $ \pm $ 1.69 & 65.27 $ \pm $ 6.83 & 66.13 $ \pm $ 9.57 & 43.29 $ \pm $ 3.10 \\
		NDFS         & 64.50 $ \pm $ 4.19 & 59.54 $ \pm $ 4.52 & 60.53 $ \pm $ 3.35 & 12.12 $ \pm $ 0.69 & 65.05 $ \pm $ 3.41 & 67.23 $ \pm $ 7.98 & 66.79 $ \pm $ 8.81 & \textbf{45.12 $ \pm $ 3.27} \\
		RUFS         & 62.48 $ \pm $ 3.97 & 60.14 $ \pm $ 4.45 & 61.28 $ \pm $ 3.35 & \textbf{14.66 $ \pm $ 0.91} & \underline{65.78 $ \pm $ 2.88} & 68.52 $ \pm $ 7.79 & 68.38 $ \pm $ 8.44 & 44.39 $ \pm $ 2.84 \\
		AUFS         & 48.77 $ \pm $ 2.69 & 57.23 $ \pm $ 4.31 & 61.82 $ \pm $ 3.14 & 11.17 $ \pm $ 0.48 & 63.37 $ \pm $ 3.11 & 66.82 $ \pm $ 6.78 & 70.14 $ \pm $ 9.62 & 44.19 $ \pm $ 2.49 \\
		MMFS\_minP    & \textbf{70.39 $ \pm $ 2.25} & 59.08 $ \pm $ 3.99 & \underline{68.51 $ \pm $ 2.83} & \underline{14.50 $ \pm $ 0.67} & 65.52 $ \pm $ 2.99 & 65.20 $ \pm $ 5.93 & 57.41 $ \pm $ 7.54 & \underline{44.47 $ \pm $ 3.26} \\
		MMFS\_maxP    & 64.87 $ \pm $ 3.35 & \textbf{65.60 $ \pm $ 2.87} & 68.20 $ \pm $ 2.28 & 9.50 $ \pm $ 0.34  & \textbf{68.13 $ \pm $ 2.70} & \textbf{81.60 $ \pm $ 6.54} & \textbf{73.20 $ \pm $ 9.98} & 42.78 $ \pm $ 3.62 \\
		MMFS\_inter   & \underline{67.20 $ \pm $ 3.09} & \underline{65.14 $ \pm $ 2.86} & \textbf{69.58 $ \pm $ 2.57} & 12.47 $ \pm $ 0.35 & 65.24 $ \pm $ 3.45 & \underline{78.70 $ \pm $ 4.26} & \underline{70.76 $ \pm $ 7.86} & 43.95 $ \pm $ 3.53\\
		\hline
	\end{tabular}
\end{table*}

\begin{table*}[t]
	\caption{NMI (NMI\%$  \pm  $STD\%) of different unsupervised feature selection methods.
		\protect\\the best result are enlighten in bold and the second best is underlined.}
	\centering
	\begin{tabular}{lllllllll}
		\hline
		Datasets & Isolet1    & COIL20     & AT\&T       & YaleB      & USPS       & ORL10P     & Lung       & TOX-171    \\
		\hline
		All Features & 74.35 $ \pm $ 1.58 & 73.71 $ \pm $ 2.44 & 80.51 $ \pm $ 1.82 & 12.98 $ \pm $ 0.80 & 60.88 $ \pm $ 0.92 & 75.82 $ \pm $ 5.19 & 57.58 $ \pm $ 5.55 & 13.65 $ \pm $ 3.31 \\
		LS           & 74.71 $ \pm $ 1.60 & 69.99 $ \pm $ 1.99 & 78.75 $ \pm $ 1.63 & 14.77 $ \pm $ 0.74 & 59.62 $ \pm $ 1.95 & 68.30 $ \pm $ 4.11 & 43.05 $ \pm $ 4.75 & 8.36 $ \pm $ 2.95  \\
		MCFS         & 74.60 $ \pm $ 1.77 & 73.35 $ \pm $ 2.35 & 80.16 $ \pm $ 1.85 & 14.16 $ \pm $ 0.91 & 60.98 $ \pm $ 1.75 & 75.54 $ \pm $ 5.24 & 45.79 $ \pm $ 4.71 & 10.18 $ \pm $ 2.01 \\
		UDFS         & 73.30 $ \pm $ 1.88 & 72.63 $ \pm $ 2.11 & 79.70 $ \pm $ 1.71 & 17.61 $ \pm $ 0.70 & 58.24 $ \pm $ 0.74 & 73.29 $ \pm $ 4.12 & 47.72 $ \pm $ 6.59 & 15.31 $ \pm $ 5.09 \\
		NDFS         & 77.86 $ \pm $ 1.66 & 73.88 $ \pm $ 2.29 & 79.80 $ \pm $ 1.66 & 19.29 $ \pm $ 0.79 & 60.62 $ \pm $ 1.39 & 75.06 $ \pm $ 4.94 & 48.35 $ \pm $ 4.94 & \underline{18.12 $ \pm $ 4.31} \\
		RUFS         & 76.70 $ \pm $ 1.75 & 73.93 $ \pm $ 2.32 & 80.54 $ \pm $ 1.70 & \underline{23.00 $ \pm $ 0.77} & 61.50 $ \pm $ 1.34 & 77.32 $ \pm $ 4.69 & 50.12 $ \pm $ 5.42 & 16.41 $ \pm $ 4.06 \\
		AUFS         & 65.67 $ \pm $ 1.50 & 72.36 $ \pm $ 2.27 & 81.02 $ \pm $ 1.66 & 17.93 $ \pm $ 1.01 & 59.60 $ \pm $ 1.18 & 77.15 $ \pm $ 4.78 & 51.78 $ \pm $ 4.85 & 15.89 $ \pm $ 3.94 \\
		MMFS\_minP    & \underline{79.32 $ \pm $ 1.1}3 & 74.38 $ \pm $ 1.54 & 84.74 $ \pm $ 1.38 &\textbf{24.60 $ \pm $ 0.68} & \underline{61.69 $ \pm $ 0.74} & 73.27 $ \pm $ 3.39 & 45.51 $ \pm $ 4.08 & \textbf{20.72 $ \pm $ 3.59} \\
		MMFS\_maxP    & 78.42 $ \pm $ 0.98 & \textbf{78.03 $ \pm $ 1.71} & \underline{85.35 $ \pm $ 0.85/} & 14.95 $ \pm $ 0.33 & \textbf{64.13 $ \pm $ 1.52} & \textbf{85.96 $ \pm $ 2.85} & \textbf{61.56 $ \pm $ 4.96} & 15.10 $ \pm $ 3.77 \\
		MMFS\_inter   & \textbf{79.49 $ \pm $ 1.30} & \underline{77.02 $ \pm $ 1.68} & \textbf{85.73 $ \pm $ 1.06} & 20.76 $ \pm $ 0.38 & 61.36 $ \pm $ 0.93 & \underline{84.34 $ \pm $ 2.20 }& \underline{60.16 $ \pm $ 3.55} & 17.55 $ \pm $ 7.11\\
		\hline
	\end{tabular}
\end{table*}

\section{Experiments}
In this section, we test the proposed feature selection method in publicly available data sets and compare our methods with several state-of-the-art methods.
\subsection{Data Sets}
In order to validate the method proposed in this paper, the experiments are conducted on 8 benchmark data sets. The details of these 8 data sets are also summarized in Table \uppercase\expandafter{\romannumeral2}.
\subsubsection{Isolet1 \cite{fanty1991spoken}}
It contains 1560 voice instaces for the name of each letter of the 26 alphabets.
\subsubsection{COIL20 \cite{nene1996columbia}}
It contains 20 objects. The images of each objects were taken 5 degrees apart as the object is rotated on a turntable and each object has 72 images. The size of each image is 32$ \times  $32 pixels, with 256 grey levels per pixel. Thus, each image is represented by a 1024-dimensional vector.
\subsubsection{AT\&T \cite{samaria1994parameterisation}}
 It contains 40 classes, and each person has 10 images. We simply use the cropped images and the size of each image is 28$ \times  $23 pixels.
\subsubsection{YaleB \cite{georghiades1998illumination}}
It contains 2414 images for 38 individuals. We simply use the cropped images and the size of each image is 32$ \times  $32 pixels.
\subsubsection{USPS \cite{hull1994database}}
 The USPS handwritten digit database. It contains 9298 16$ \times  $16 handwritten images of ten digits in total.
\subsubsection{ORL10P \cite{samaria1994parameterisation}}
It contains 100 instances for 10 classes. And each image is represented by a 10304-dimensional vector.
\subsubsection{Lung \cite{bhattacharjee2001classification}}
It contains 5 classes and 203 instances in total, where each instance consists of 3312 features.
\subsubsection{TOX-171 \cite{stienstra2010kupffer}}
It contains 171 instances in four categories and each instance has 5748 features.

\subsection{Experimental Setting}
In order to prove the efficiency of the proposed approach, on each data set, we compare the proposed algorithms with the following six unsupervised feature selection methods.
\subsubsection{Baseline}
All features.
\subsubsection{Laplacian Score (LS) \cite{he2006laplacian}}
It evaluated the importance of features according to the largest Laplacian scores, which used to measure the power of locality preservation.

\subsubsection{Multi-Cluster Feature Selection (MCFS) \cite{cai2010unsupervised}}
According to the spectral analysis of the data and $L_1$-regularized models for subset selection, it select those features that the multi-cluster structure of the data can be best preserved.
\subsubsection{Unsupervised Discriminative Feature Selection (UDFS) \cite{yang2011l2}}
It selects features through a joint framework that combines the discriminative analysis and $l_{2,1}$-norm-norm minimization.

\subsubsection{Nonnegative Discriminative Feature Selection (NDFS) \cite{li2012unsupervised}}
In order to select the most discriminative features, NDFS performs spectral clustering to learn the cluster labels of the input samples and adds $l_{2,1}$-norm minimization constraint to reduce the redundant or even noisy features.

\subsubsection{Robust Unsupervised Feature Selection (RUFS) \cite{qian2013robust}}
It selects features by a joint framework of local learning regularized robust nonnegative matrix factorization and $l_{2,1}$-norm minimization.
\subsubsection{Adaptive Unsupervised Feature Selection (AUFS) \cite{qian2015joint}}
It uses a joint adaptive loss for data fitting and a $l_{2,0}$ minimization for feature selection.

\par For LS, MCFS, UDFS, NDFS, RUFS, AUFS, and MMFS, we fixed the number of neighbors as 5 for all data sets. As for regularization parameter $ \lambda $  in problem (9), the optimal parameter is selected at the candidate set \{0.001, 0.01, 0.1, 1, 10, 100, 1000\}, and parameter $n$ is selected at the set \{5, 6, 7, ..., 18, 19, 20\}. In order to make a fair comparison between different unsupervised feature selection methods, we fix the number of the selected features as {50, 100, ..., 250, 300} for all data sets except the USPS data set. For USPS, the number of the selected features was set to {50, 80, ..., 170, 200}, because its  total number of features is 256.

\begin{figure*}	
	\centering
	\subfigure[]{
		\label{Fig.sub.a}
		\includegraphics[width=3.9cm]{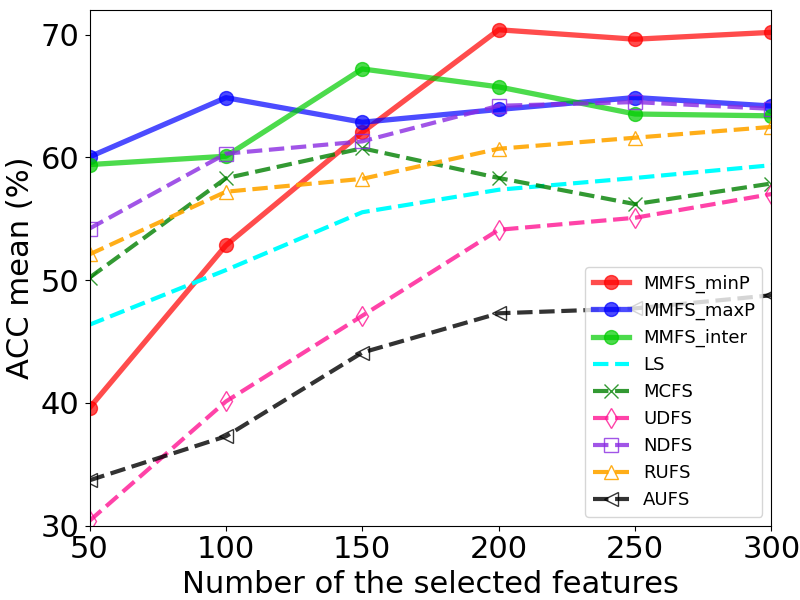}}
	\subfigure[]{
		\label{Fig.sub.b}
		\includegraphics[width=3.9cm]{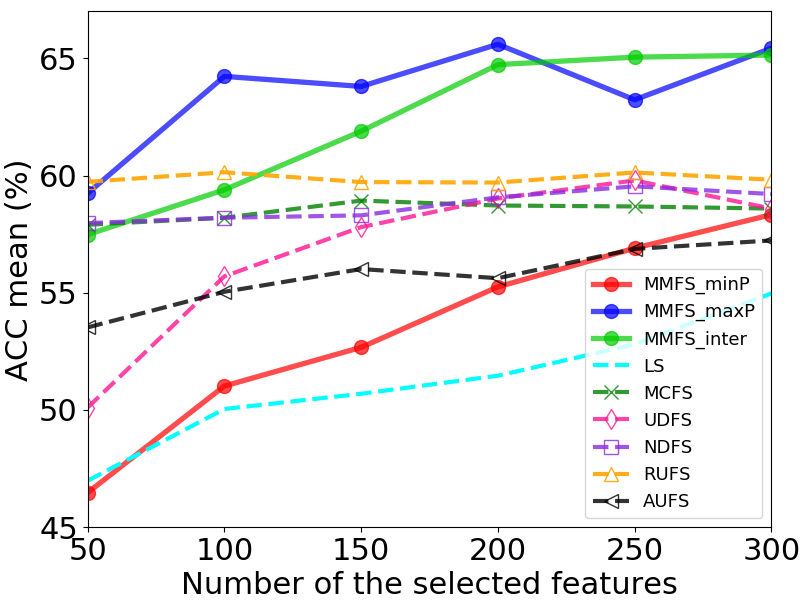}}
	\subfigure[]{
		\label{Fig.sub.c}
		\includegraphics[width=3.9cm]{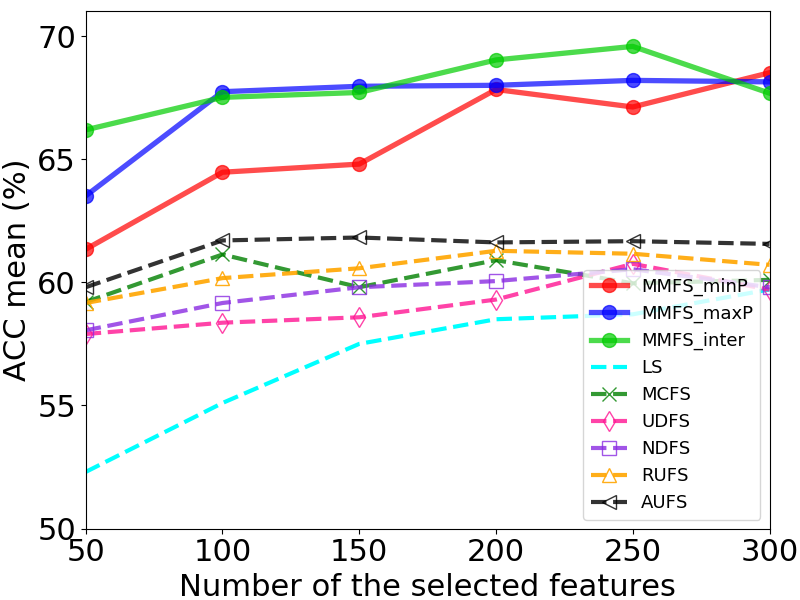}}
	\subfigure[]{
		\label{Fig.sub.d}
		\includegraphics[width=3.9cm]{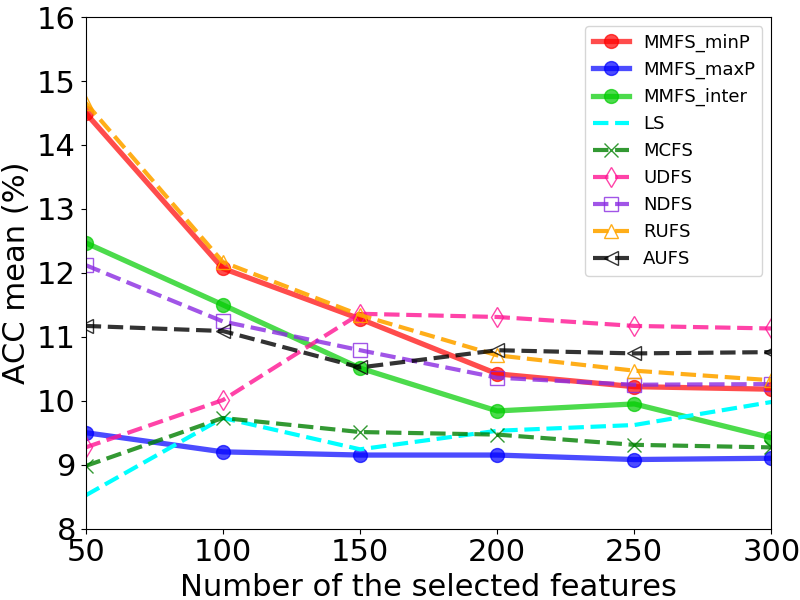}}
	\subfigure[]{
		\label{Fig.sub.e}
		\includegraphics[width=3.9cm]{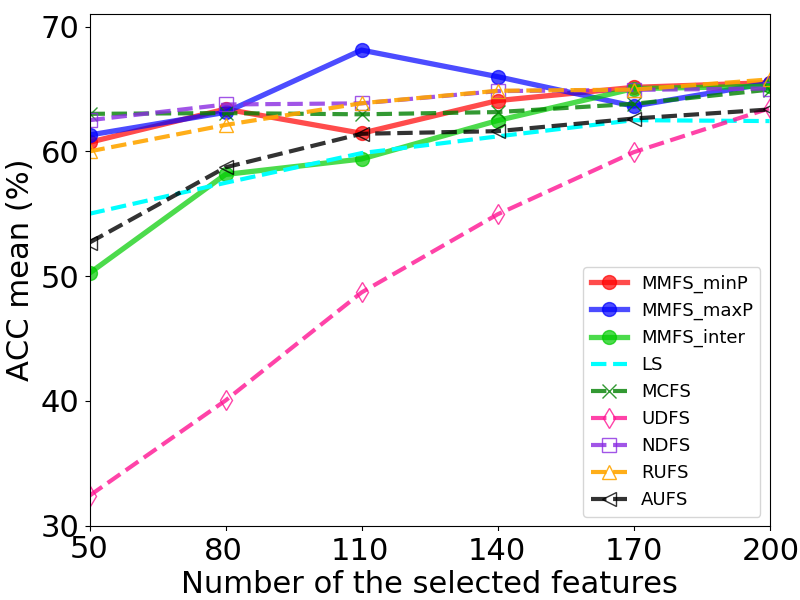}}
	\subfigure[]{
		\label{Fig.sub.f}
		\includegraphics[width=3.9cm]{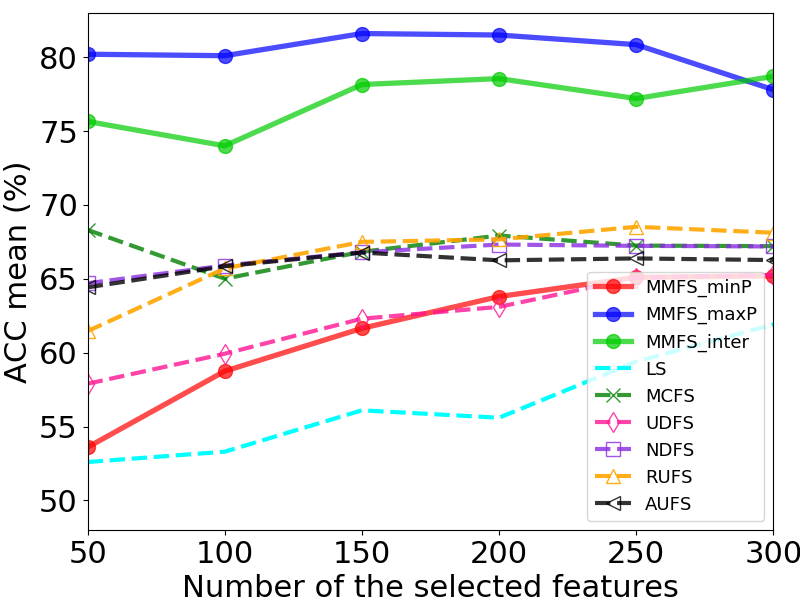}}
	\subfigure[]{
		\label{Fig.sub.g}
		\includegraphics[width=3.9cm]{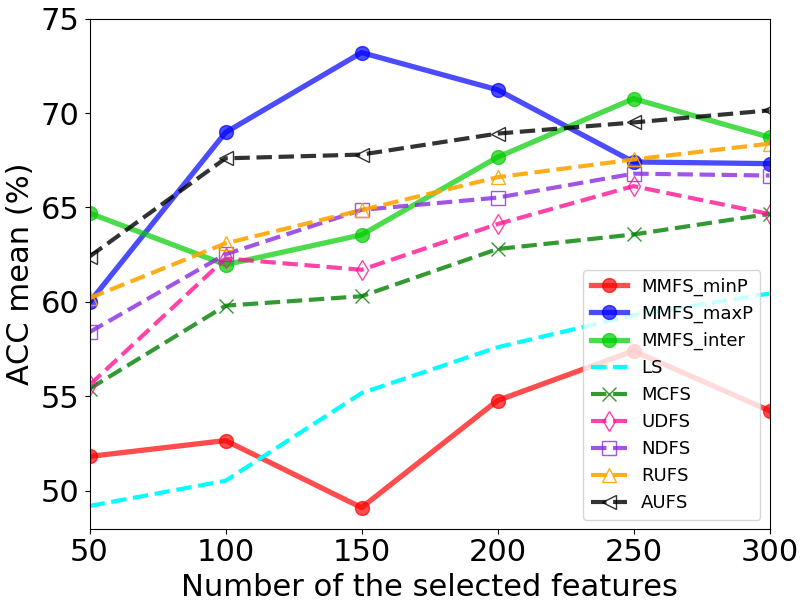}}
	\subfigure[]{
		\label{Fig.sub.h}
		\includegraphics[width=3.9cm]{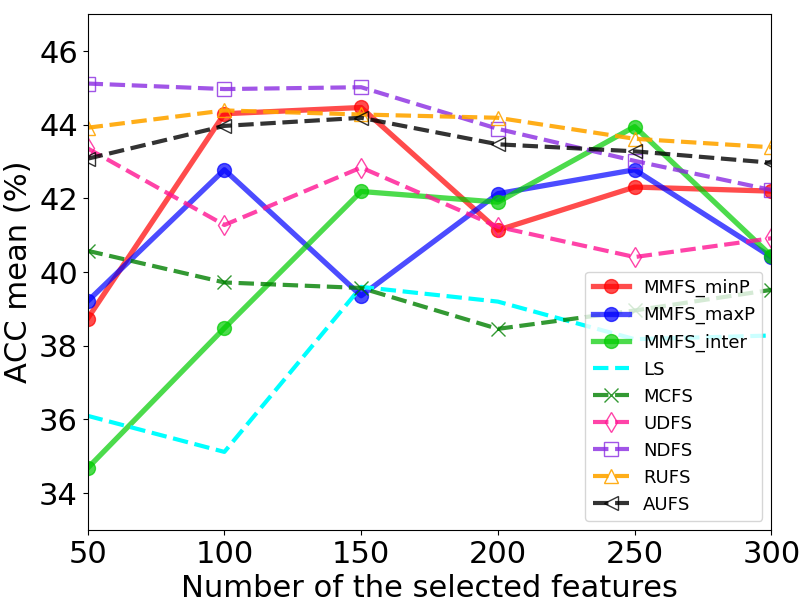}}
	\caption{ACC of various feature selection methods with different number of selected features.
		(a) Isolet1. (b) COIL20. (c) AT\&T. (d) YaleB. (e) USPS. (f) ORL10P. (g) Lung. (h)TOX-171}
	\label{Fig.main}
\end{figure*}
\begin{figure*}[t]
	\centering
	\subfigure[]{
		\label{Fig.sub.a}
		\includegraphics[width=3.9cm]{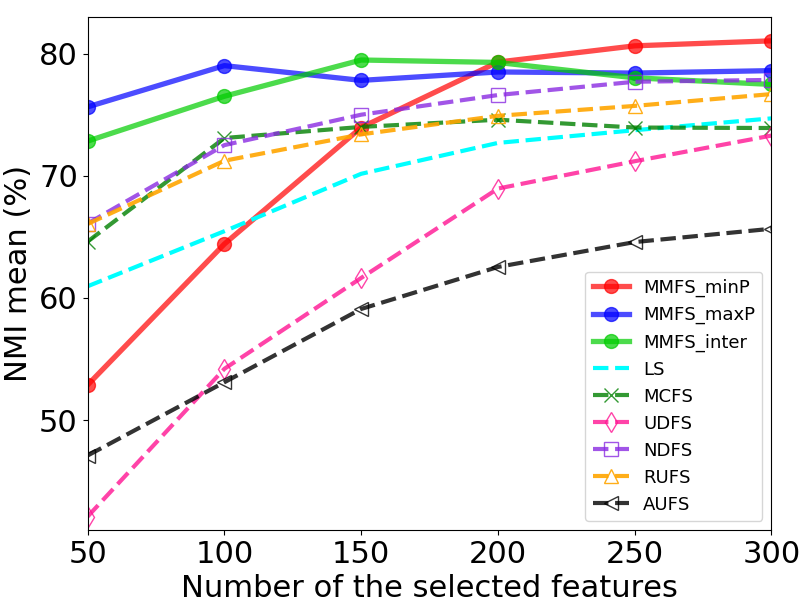}}
	\subfigure[]{
		\label{Fig.sub.b}
		\includegraphics[width=3.9cm]{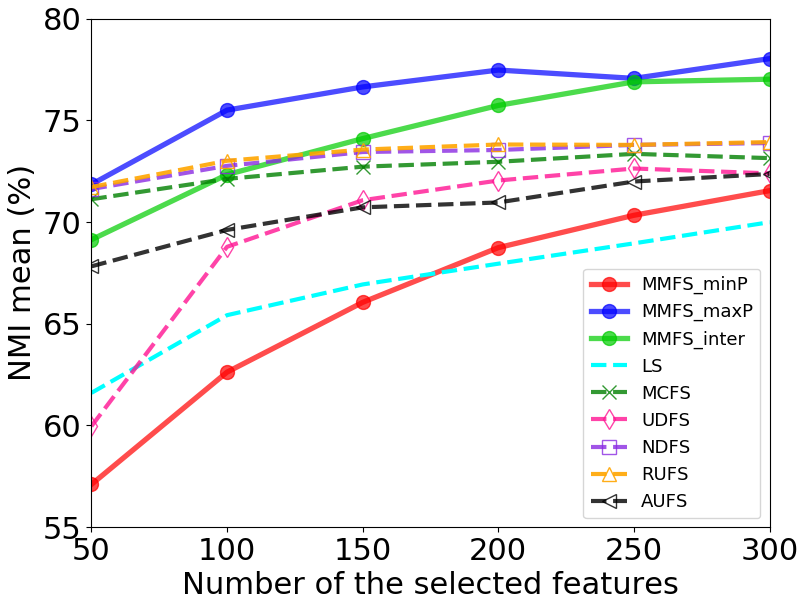}}
	\subfigure[]{
		\label{Fig.sub.c}
		\includegraphics[width=3.9cm]{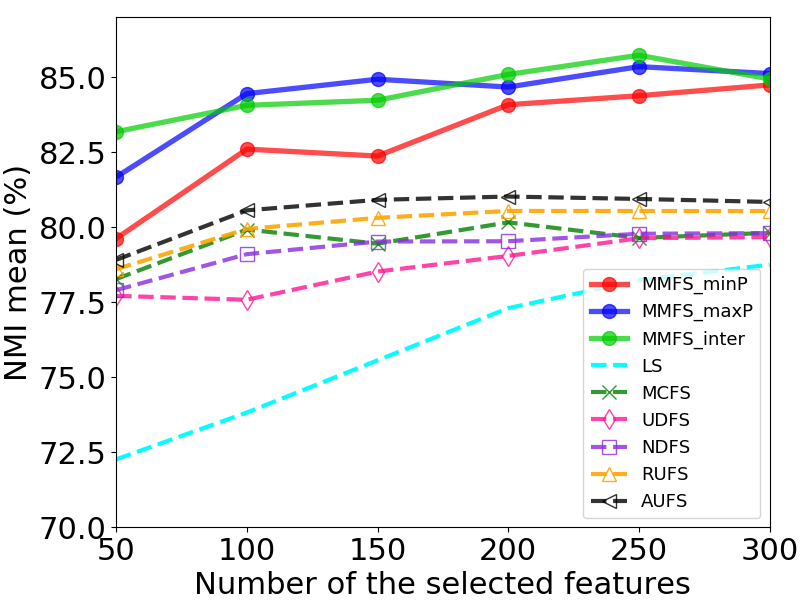}}
	\subfigure[]{
		\label{Fig.sub.d}
		\includegraphics[width=3.9cm]{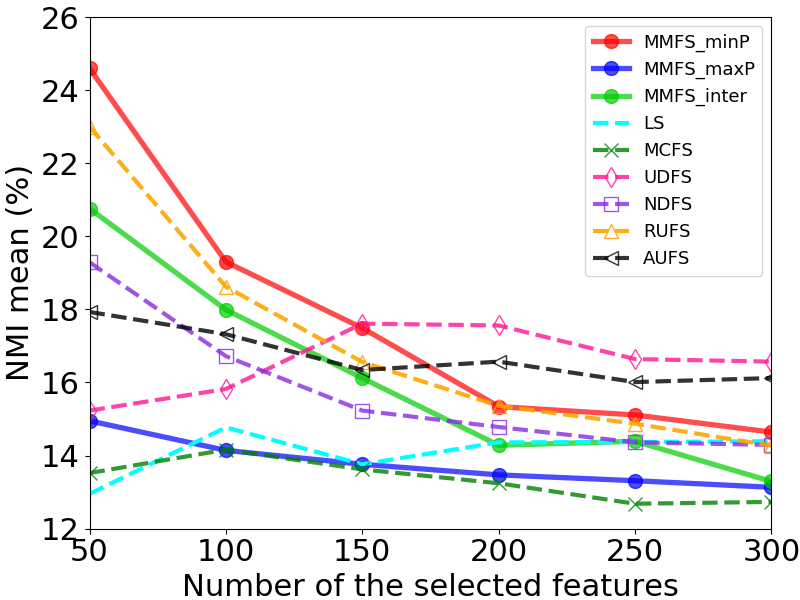}}
	\subfigure[]{
		\label{Fig.sub.e}
		\includegraphics[width=3.9cm]{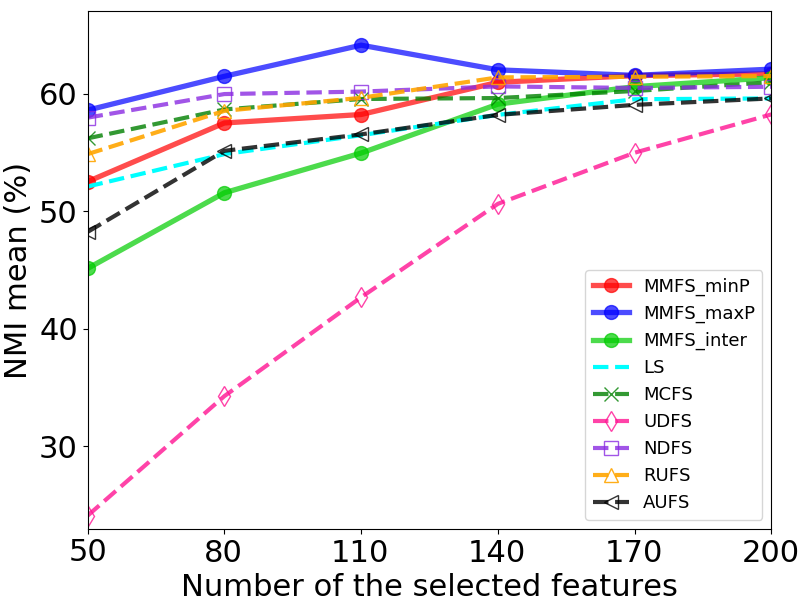}}
	\subfigure[]{
		\label{Fig.sub.f}
		\includegraphics[width=3.9cm]{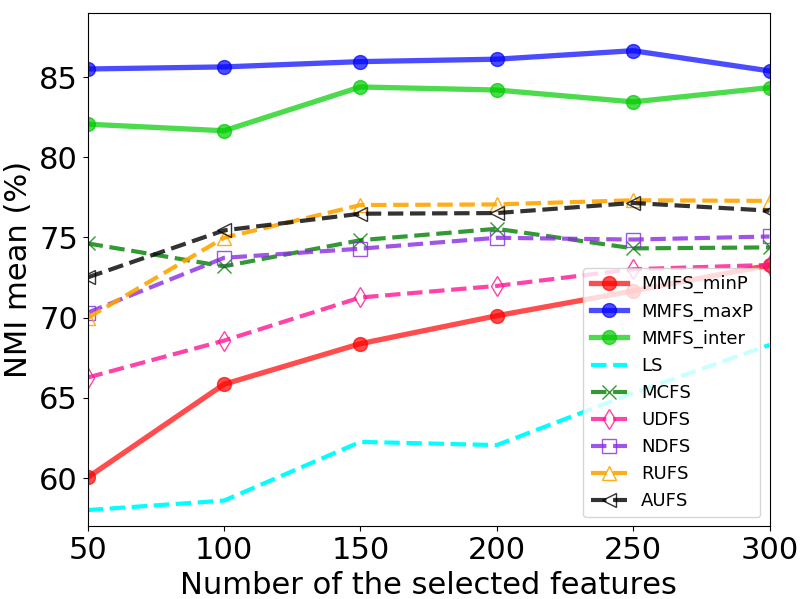}}
	\subfigure[]{
		\label{Fig.sub.g}
		\includegraphics[width=3.9cm]{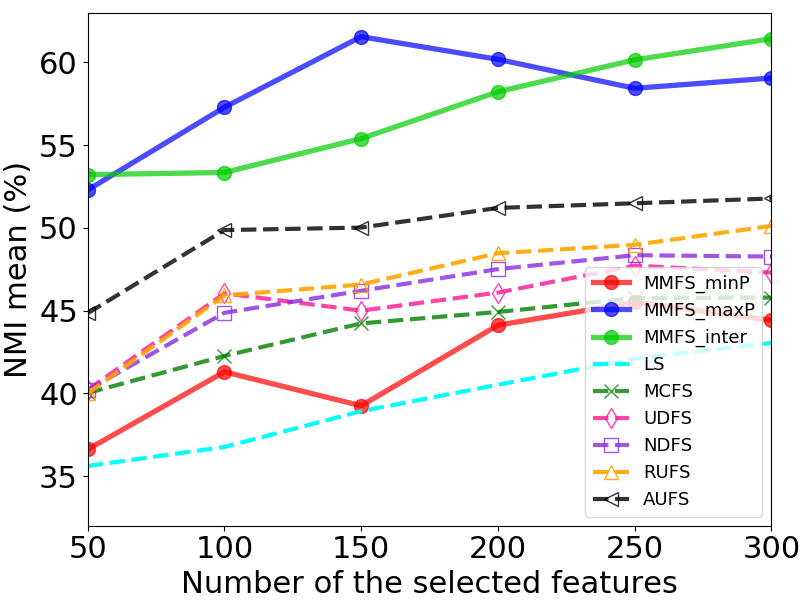}}
	\subfigure[]{
		\label{Fig.sub.h}
		\includegraphics[width=3.9cm]{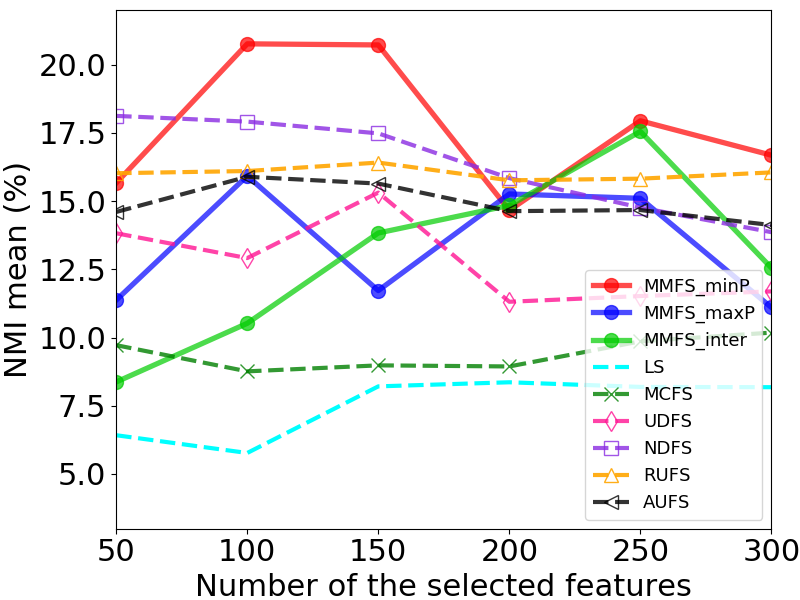}}
	\caption{NMI of various feature selection methods with different number of selected features.
		(a) Isolet1. (b) COIL20. (c) AT\&T. (d) YaleB. (e) USPS. (f) ORL10P. (g) Lung. (h)TOX-171}
	\label{Fig.main}
\end{figure*}

\par 
We use K-means clustering algorithm to perform the clustering task. Since K-means clustering is sensitive to its initialization of the clustering seeds, we repeat the experiments 20 times with random initialization of the seeds, and record the mean and standard deviation of ACC and NMI.
Given a data set, let $ {p_i}$ and $ {q_i}$  be the labels obtained by clustering algorithm and the labels provided by data set respectively. And the ACC  \cite{wang2016sparse} can be defined as follows:
\begin{equation}
ACC = \frac{{\sum\nolimits_{i = 1}^n {\delta \left( {{p_i},\mbox{map}\left( {{q_i}} \right)} \right)} }}{n}
\end{equation}
where if $a = b$, then $ \delta (a, b) = 1$, otherwise $ \delta (a, b) = 0$. The map($\cdot$) represents an optimal permutation mapping function that matches the clustering label obtained by clustering algorithm with the label of ground-truth. The construction of mapping function can refer to  the Kuhn-Munkres algorithm \cite{lovasz2009matching}. The higher the ACC is, the better the clustering performance is.

Normalized Mutual Information (NMI) is an normalization of the Mutual Information (MI) score to scale the results between 0 and 1. NMI is the similarity between the clustering results and the ground-truth labels of the data sets. The higher the NMI is, the better the clustering performance is. Scikit-learn (sklearn) \cite{scikit-learn} is a third-party module in machine learning, which encapsulates the common machine learning methods. In our experiment, we calculate NMI by a method in package sklearn as follows:

sklearn.metrics.normalized\_mutual\_info\_score (labels\_true, labels\_pred).

\subsection{Experimental Results and Analysis}

We conclude the experimental results in Table \uppercase\expandafter{\romannumeral3} and \uppercase\expandafter{\romannumeral4}. And the clustering accuracy and NMI with different number of features are shown in Fig. 2 and Fig.3. The results of the compared methods in the figures and tables refer to some experimental data from the previous method \cite{han2018unified}, while our experimental settings are the same as \cite{han2018unified}, such as the number of neighboring parameter, the number of repetition of the experiment and the number of selected features. The higher the mean is and the smaller the standard deviation is, the better the clustering performance is.
\begin{figure*}	
	\centering
	\subfigure[]{
		\label{Fig.sub.a}
		\includegraphics[width=5.3cm]{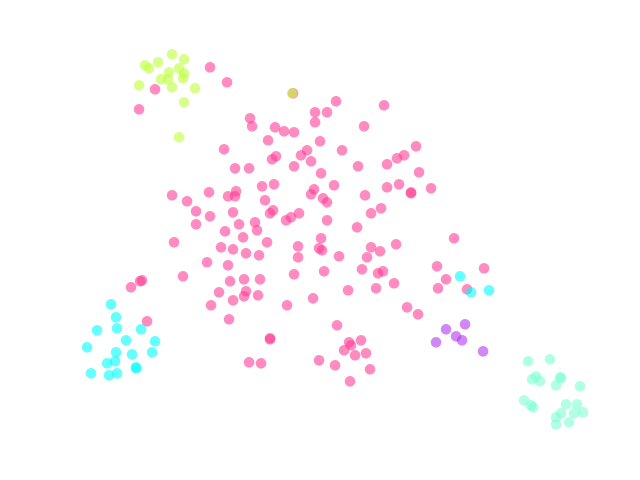}}
	\subfigure[]{
		\label{Fig.sub.b}
		\includegraphics[width=5.3cm]{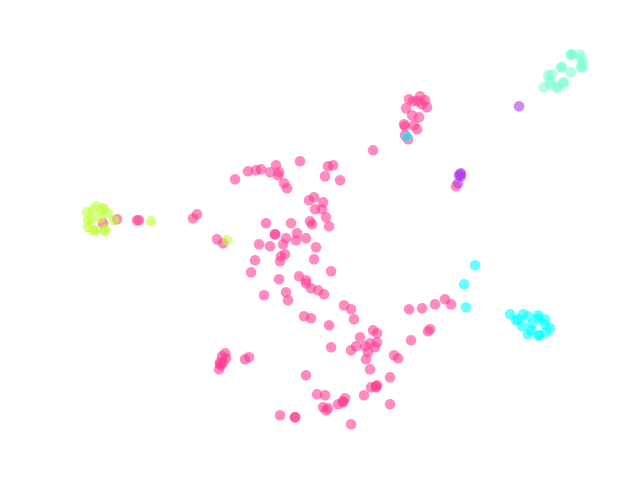}}
	\subfigure[]{
		\label{Fig.sub.c}
		\includegraphics[width=5.3cm]{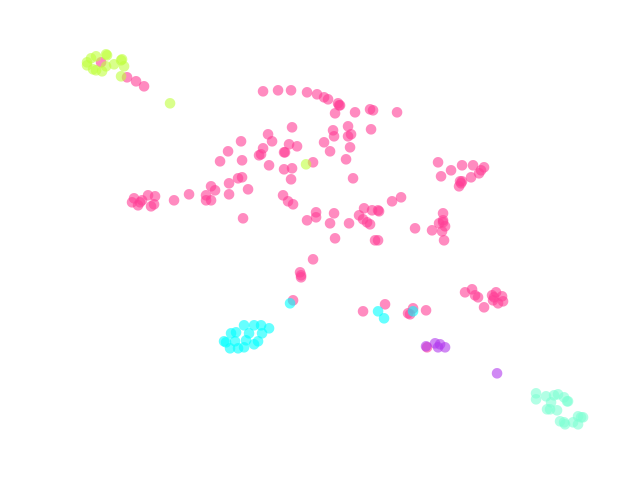}}
	\caption{The distribution of Lung data after using different algorithms.
		(a) Initial state (b) MMFS\_minP. (c) MMFS\_maxP.}
	\label{Fig.main}
\end{figure*}

\par According to Tables \uppercase\expandafter{\romannumeral3} and \uppercase\expandafter{\romannumeral4}, it can be observed that our methods are significantly better than all other methods on two data sets, i.e., AT\&T, and Isolet1 data sets. On other data sets, the performances of MMFS\_minP and MMFS\_maxP are different. In general, when the original data distribution is closer to the ideal situation (most of the data in the same category distributed together), the performance of MMFS\_maxP will be better, such as COIL20, USPS, ORL10P and Lung. Otherwise, when the data distribution is more chaotic, the performance of MMFS\_minP will be better, such as Isolet1, YaleB, TOX-171. The reason is that these two algorithms of MMFS keep the original data manifold structure in different ways accordingly; they retain a compact or loose structure. Our algorithms do not perform very well on YaleB and TOX-171 data sets; one possible reason is that our methods cannot adapt to all kinds of data distributions, so they cannot retain more accurate data structure information.
\par As we can see from Figs. 2 and 3, the proposed algorithms MMFS\_minP,  MMFS\_maxP and MMFS\_inter have better performance than other methods on most of data sets. However, in some data sets, such as Isolet1 and COIL20, the performance of MMFS\_minP is not effective enough when the number of selected features is not large enough.  The performance of MMFS\_inter is usually between that of MMFS\_minP and MMFS\_maxP except USPS, a possible reason is that the way we combine the two methods is not flexible enough so that we can not get a better feature subset. At the same time, we also observe that feature selection technology improves the clustering performance. For example, when the ratio of the number of features used to the total number of features is very small, but its accuracy is much higher than that of using all features in most data sets. Even in the datasets YaleB and Tox-171, ACC with fewer features is also much better than that observed with all features.

In most cases, the structural information is helpful to select the more effective features. MCFS, UDFS, NDFS, RUFS and AUFS all apply local structure information. However, the algorithms we proposed not only retain the local structural information, but also obtain the association information between non-adjacent points, and make more sufficient use of the obtained structure information, which is an important reason for our algorithms to obtain good results.

Fig. 4 presents the two-dimensional data distribution after adding the relations obtained by the algorithms MMFS\_minP and MMFS\_maxP  on the Lung data set. The dimension reduction method used for visualization  refers to t-SNE method \cite{maaten2008visualizing}. The first part of Fig. 4 is obtained by projecting data matrix \textbf{X} into a two-dimensional plane, which depicts the original two-dimensional data distribution of the Lung data set. The second and third parts of Fig. 4 are obtained by projecting $ \textbf{X}^T\textbf{W} $ into a two-dimensional plane, where \textbf{W} is the optimal weight matrix obtained by MMFS\_minP and MMFS\_maxP respectively. Therefore, the second and third parts respectively describe the two-dimensional data distribution of the Lung data set under the effect of the relationship obtained by the algorithms MMFS\_minP and MMFS\_maxP. It is observed that some data points in the two-dimensional projection are obviously shortened or even overlapped by the influence of the relationship. Thus, the proposed algorithms MMFS\_minP and MMFS\_maxP can get the relationship of data that makes the data points in the same class more closer.

\begin{figure}[t]	
	\centering
	\subfigure[]{
		\label{Fig.sub.a}
		\includegraphics[width=3.9cm]{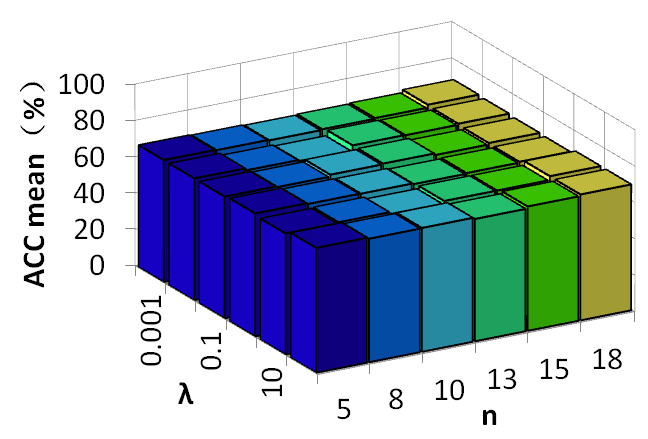}}
	\subfigure[]{
		\label{Fig.sub.b}
			\includegraphics[width=3.9cm]{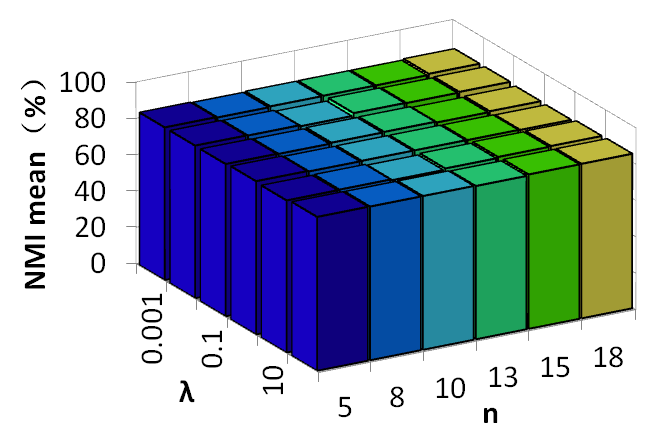}}
	\subfigure[]{
		\label{Fig.sub.c}
		\includegraphics[width=3.9cm]{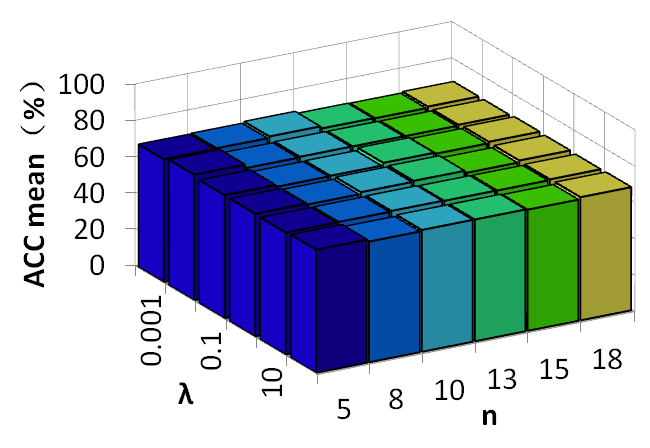}}
	\subfigure[]{
		\label{Fig.sub.d}
		\includegraphics[width=3.9cm]{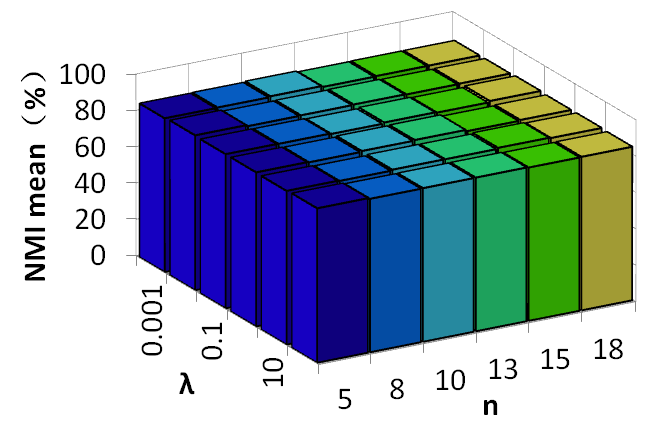}}
	\subfigure[]{
		\label{Fig.sub.e}
		\includegraphics[width=3.9cm]{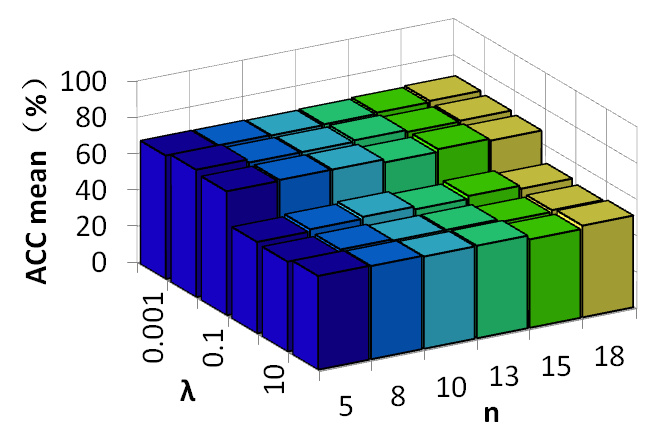}}
	\subfigure[]{
		\label{Fig.sub.f}
		\includegraphics[width=3.9cm]{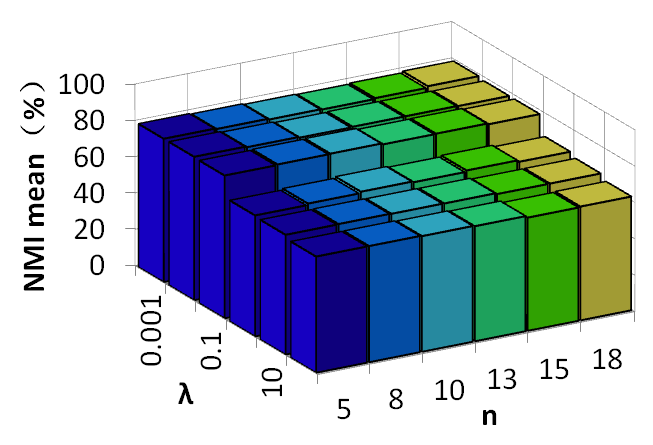}}
	\subfigure[]{
		\label{Fig.sub.g}
		\includegraphics[width=3.9cm]{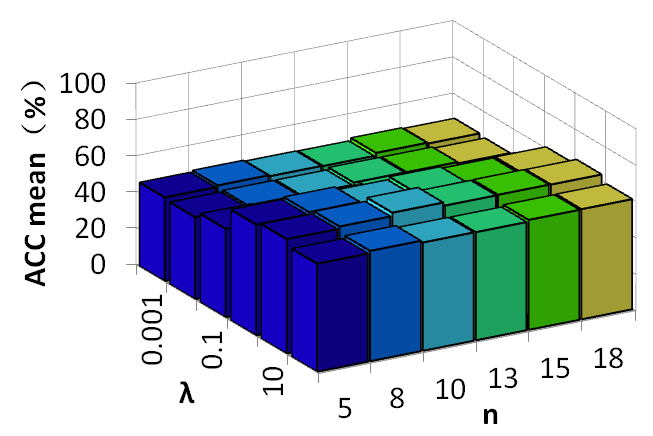}}
	\subfigure[]{
		\label{Fig.sub.h}
		\includegraphics[width=3.9cm]{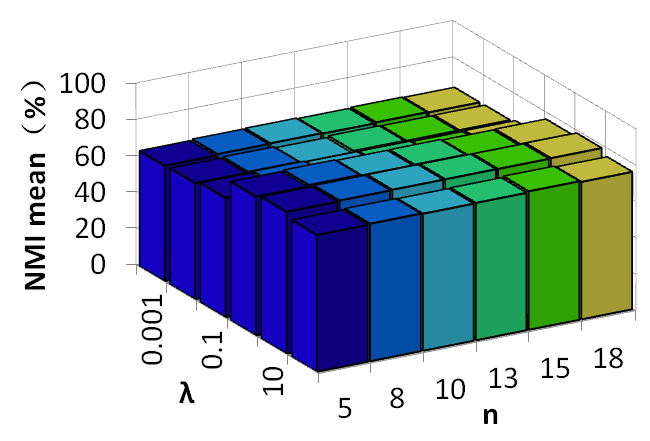}}
	\caption{Parameter sensitivity demonstration on different data sets on top 200 features. (a)-(d) AT\&T. (e)-(h) Isolet1. (a), (b), (e) and (f) MMFS\_minP. (c), (d), (g) and (h) MMFS\_maxP.}
	\label{Fig.main}
\end{figure}

\subsection{Parameter Sensitivity}
We use AT\&T and Isolet1 data sets to measure the sensitivity of the algorithms MMFS\_minP and MMFS\_maxP to parameters ($ \lambda $ and $n$), and the results (ACC mean and NMI mean) are shown in Fig. 5.
It shows the performances of MMFS\_minP (the four subplots on the left, (a), (b), (e) and (f)) and MMFS\_maxP (the four subplots on the right, (c), (d), (g) and (h)) on AT\&T (the first four subplots, (a)-(d)) and Isolet1 (the last four subplots, (e)-(h)) data sets respectively when the parameters $ \lambda $ and $n$ take different values.

From the first four subplots, we can observe that our methods are not significantly sensitive to the regularization parameter $ \lambda $ and the number of step $n$ for the data set AT\&T. However, for the data set Isolet1, our methods are more sensitive to $ \lambda $ than $n$. For example, In the subfigure (e) of Fig. 5, when the parameter $n$ is fixed, the change of $ \lambda $ has a significant impact on the performance. The reason is that  $ \lambda $  controls the sparsity of \textbf{W}. On the other hand, when the parameter $ \lambda $ is fixed, the change of $n$ has little impact on the performance. This confirm that the proposed new parameter $n$ is rather robust to our algorithms. For the selection  of the parameter $\lambda $, we follow the traditional way \cite{han2018unified}, i.e., a grid search strategy from the candidate set is used to select the best parameter.

\section{Conclusion}
We proposed a new feature selection approach, MMFS, which can preserve the manifold structure of high dimensional data. There are two ways to achieve our purpose, MMFS\_minP and MMFS\_maxP, and we also combine these two algorithms into an algorithm MMFS\_inter. The new framework learns a weight matrix $\mathbf{W}$ which projects the data to close the data structure constructed by multi-step Markov transition probability; $l_{2,1}$-norm is applied  to make $\mathbf{W}$ to be row sparse for feature selection. An iterative optimization algorithm is proposed to optimize the new model. We perform comprehensive experiments on eight public data sets to validate the effectiveness of the proposed approach.


%

\appendices




\ifCLASSOPTIONcaptionsoff
  \newpage
\fi



%
\bibliographystyle{IEEEtran}
\bibliography{BibFile}



%








\end{document}